\pdfoutput=1
\documentclass[11pt]{article}
\usepackage{acl}
\usepackage{times}
\usepackage{latexsym}
\usepackage[T1]{fontenc}
\usepackage[utf8]{inputenc}

\usepackage{microtype}
\usepackage{hyperref}
\usepackage{url}
\usepackage{booktabs}
\usepackage{wrapfig}

\usepackage{microtype}

\usepackage{inconsolata}

\usepackage{graphicx}

\usepackage{multirow}
\usepackage{amsmath}
\usepackage{tabularray}
\usepackage[T1]{fontenc}
\usepackage{enumitem}
\usepackage{multirow}
\usepackage{color, colortbl}
\definecolor{LightCyan}{rgb}{0.94,1,1}
\definecolor{Red2}{rgb}{1,0.85,0.85}
\definecolor{Red1}{rgb}{1,0.7,0.7}
\definecolor{Blue}{rgb}{0.8,0.95,1}
\definecolor{Gray}{rgb}{0.8,0.8,0.8}
\definecolor{LightGray}{rgb}{0.91,0.91,0.91}
\definecolor{LLightGray}{rgb}{0.97,0.97,0.97}
\definecolor{mygreen}{rgb}{0.4,0.7,0.306}
\definecolor{myblue}{rgb}{0.294,0.447,0.796}
\usepackage{hhline}
\usepackage{adjustbox}
\usepackage{float}
\usepackage{dblfloatfix}
\usepackage{xcolor}
\usepackage[hang,flushmargin]{footmisc}

\usepackage{arydshln}
\usepackage{color, colortbl}

\usepackage[textsize=tiny]{todonotes}

\title{TextLap: Customizing Language Models for Text-to-Layout Planning}

\author{
    Jian Chen\textsuperscript{1}\thanks{Work done at University at Buffalo.},
    Ruiyi Zhang\textsuperscript{2}\thanks{Corresponding Author},
    Yufan Zhou\textsuperscript{2}, Jennifer Healey\textsuperscript{2},
     \\
    \textbf{Jiuxiang Gu\textsuperscript{2}, Zhiqiang Xu\textsuperscript{3},
    Changyou Chen\textsuperscript{1}}
    \\
    \\
    \vspace{-2em}
    \textsuperscript{1}University at Buffalo~~~~~~~~
    \textsuperscript{2}Adobe Research~~~~~~~~
    \textsuperscript{3}MBZUAI
}

\IfFileExists{main_backup.aux}{
  \message{We saw a default backup.aux file, let's use it instead of the main aux file.}
  \nofiles % Disable default aux file
  \makeatletter
  \input{main_backup.aux}
  \makeatother
}{}

\begin{document}
\maketitle

\begin{abstract}
Automatic generation of graphical layouts is crucial for many real-world applications, including designing posters, flyers, advertisements, and graphical user interfaces. Given the incredible ability of Large language models (LLMs) in both natural language understanding and generation, we believe that we could customize an LLM to help people create compelling graphical layouts starting with only text instructions from the user. We call our method TextLap (text-based layout planning) \footnote{Data and code are available at: \\ \textcolor{magenta}{\href{https://github.com/puar-playground/TextLap}{\small https://github.com/puar-playground/TextLap}}}.  It uses a curated instruction-based layout planning dataset (InsLap) to customize LLMs as a graphic designer. We demonstrate the effectiveness of TextLap and show that it outperforms strong baselines, including GPT-4 based methods, for image generation and graphical design benchmarks. 

\end{abstract}

\begin{figure*}[t]
    \centering
    \includegraphics[width=1\textwidth]{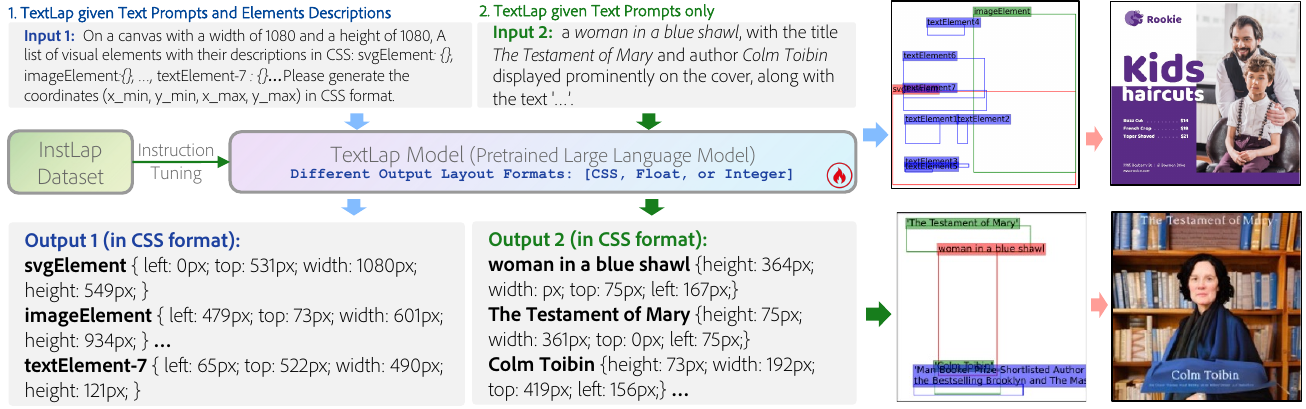}
    \vspace{-1.5em}
    \caption{Overview of TextLap fine-tuned on InstLap. 1) TextLap can perform graphic designs and output coordinates  given a list of elements including images, texts, and scalable vector graphics (SVG). The image is rendered accordingly. 2) TextLap can extract key elements from text prompts and provie their coordinates. The image can be rendered with image generation tools.}
    \label{fig:TextLap_overview}
    \vspace{-1.4em}
\end{figure*}

\section{Introduction}
Traditional layout generation tasks typically involve organizing text and graphical elements into a pleasing and professional looking arrangement.  This is often time-consuming and difficult for users, especially those without training or design skill. Our goal is to condition an LLM so that it allows users to generate professional looking layouts by simply inputting text instructions.  We are inspired by the rapid advancement of generative models for text ~\citep{ouyang2022training, achiam2023gpt}, images~\citep{betker2023improving,podell2023sdxl}, and videos~\citep{peebles2023scalable} and recent works that create layouts using graphical elements such as category, position, or size as input ~\citep{li2020attribute,kikuchi2021constrained,jyothi2019layoutvae,Inoue_2023_CVPR}. We advance the field by customizing LLMs as text-guided layout generation models that require only a text description of the desired layout and optionally text descriptions for visual elements as input, enabling layout planning solely in the text modality. We believe that building an LLM-based layout generation offers a more user-friendly approach to achieving desired designs, allowing text instructions to guide the process—an aspect that is challenging to incorporate in traditional settings.

Creating 2D graphical layouts from text alone is challenging for several reasons. First, text descriptions of 2D arrangements are usually underspecified. There are multiple ways to describe a 2D layout, and there are multiple 2D layouts that can meet a description. Furthermore, comprehending 2D spatial relations \citep{xu2020layoutlm} is challenging for LLMs due to their autoregressive nature and 1D positional embedding \citep{vaswani2017attention}. Even understanding the meaning of coordinate numbers in a description is sometimes difficult for LLMs. 

To enable the generation of 2D designs from text, we build an instruction-based layout planning (InstLap) dataset to customize large language models. We then use InstLap to train our Text-based Layout Planning (TextLap) model through supervised fine-tuning of an LLM, enabling it to understand 2D spatial relationships and generate or modify bounding box coordinates based on text prompts. Figure \ref{fig:TextLap_overview} illustrates example inputs and outputs of the proposed TextLap model.

As an LLM, TextLap enables users to iteratively refine layout designs through natural language conversations. Initial text instructions provide an initial draft, and users can further customize their designs by interacting with the model. TextLap generates text-coherent layouts in response to users’ iterative requests, enabling them to quickly find the most suitable custom template for their design. We believe that TextLap significantly reduces the time required to create designs, enhances design efficiency, and plays an important role in assisting graphic designers. Moreover, TextLap also serves as a layout planning component for image generators, greatly enhancing the layout coherence of generated images according to text prompts. Our contributions are threefold:

\begin{itemize}[topsep=0.5pt,nosep,leftmargin=*]
\item We proposed a novel text-to-layout tasks aiming to enhance design efficiency and designed evaluation dataset and protocals.
\item We built an instruction tuning dataset named InstLap with human machine hybird annotations, enabling LLMs to perform text-to-layout planning.
\item We trained a LLM-based layout planning model TextLap, and its empirical evaluations on benchmark datasets show its superior performance compared to GPT-4 models.
\end{itemize}

\section{Related Work}
\paragraph{Layout Generation Models}
Various approaches have been developed for controllable layout generation. \citet{jiang2023layoutformer++, arroyo2021variational, gupta2021layouttransformer} utilize transformers to create auto-regressive models that produce layouts as a sequence of element attributes. Adversarial generative models \cite{kikuchi2021constrained, li2020attribute} and diffusion models \cite{chen2024towards, Inoue_2023_CVPR,zhang2023layoutdiffusion} have been used to enhance the quality of generation and unify various conditional generation tasks in a single model. TextDiffuser \cite{chen2023textdiffuser} applies a layout transformer to generate layouts for keywords in text-rich images. \cite{jin2022text2poster, hsu2023posterlayout, yu2022layoutdetr} generate background-aware layouts for posters by detecting smooth regions of the background image. \cite{yang2022scene} introduces a graph transformer architecture to generate layouts for images based on a scene graph that describes the relationship between visual objects. The visual layout serves as a crucial spatial control in image generation. GLIGEN \cite{li2023gligen} integrates gated self-attention to enable spatial grounding capabilities in pre-trained diffusion models. Similarly, InstanceDiffusion \cite{wang2024instancediffusion} integrates locations and descriptions at the instance level into the generation process. TextLap is an LLM-based layout planning model that can be used to design various items, such as posters and book covers. 

\vspace{-1em}
\paragraph{LLM-based Layout Generation}
Recent studies have introduced approaches based on language models to enable interactive layout generation and modification through chatting. TextDiffuser-2 \citep{chen2023textdiffuser2} fine-tunes a large language model to generate layouts for keywords for text rendering. \cite{yang2024if} demonstrate that the integration of programming code in LLM training benefits the performance of LLM agents in various tasks. LayoutNUWA \citep{tang2023layoutnuwa} leverages a large language model to generate layouts as HTML code. LayoutGPT \citep{feng2023layoutgpt} in-context visual demonstrations in CSS structures to enhance the visual planning skills of GPT-3.5/4 \citep{ouyang2022training, achiam2023gpt} to plan layouts from text conditions. MuLan \cite{li2024mulan} iteratively plans the layout of an image by breaking down the text prompt into a sequence of sub-tasks with an LLM, and then revises the image at each step based on feedback from a vision-language model (VLM). InstructScene \citep{lin2024instructscene} uses ChatGPT to filter the caption describing the layout elements. TextLap sets itself apart from prior research by delving into the text-to-layout task, which allows users to generate a layout design based on natural-language descriptions.

\begin{figure*}[t]
    \centering
    \includegraphics[width=1\textwidth]{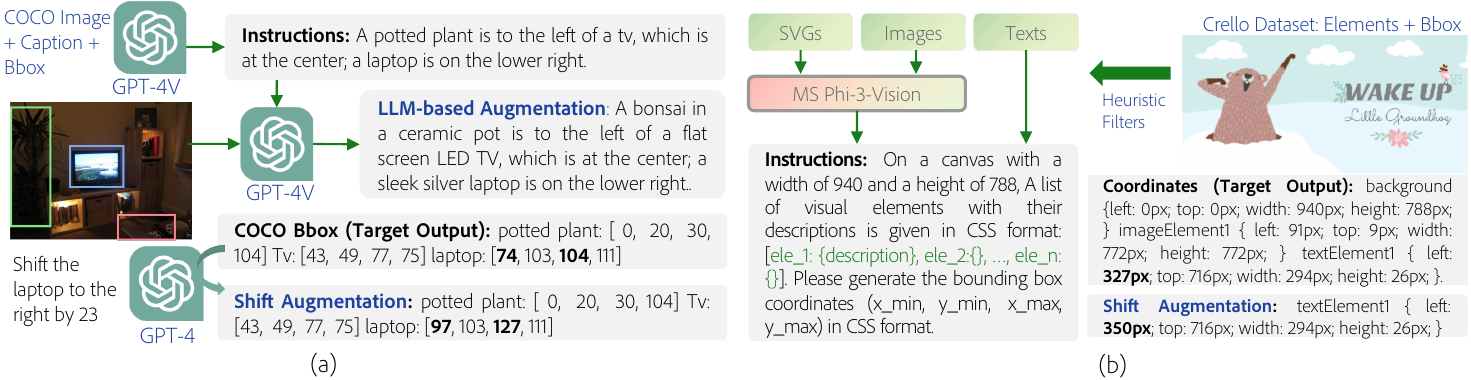}
    \vspace{-2.em}
    \caption{Overview of how to build the InstLap dataset. (a) shows how to build InstLap based on COCO dataset, which is composed of two data augmentations for input instructions and output layouts, respectively. (b) presents the how to incorporate Crello dataset into Instlap, where visual elements are first described by \texttt{Phi-3-Vision} and augmented into the text instructions.}
    \vspace{-1em}
    \label{fig:InstLap_overview}
\end{figure*}

\section{Creating the InstLap Dataset} 
\label{sec:data}
To customize a language model for layout planning, we developed the Instruction to Layout Planning (InstLap) dataset which comprises filtered and enhanced image-caption pairs from MS-COCO (2017)~\citep{lin2014microsoft}, a subset of the LAION datset~\citep{schuhmann2022laion,chen2023textdiffuser}, and Crelllo~\citep{yamaguchi2021canvasvae}.

Table \ref{tab:data_statistics} provides the statistics for the InstLap dataset, which includes human-annotated benchmarks, the original dataset, and augmented  samples. 
\begin{table}[!ht]
\fontsize{6.5}{8.5}\selectfont
\begin{adjustbox}{width=1\linewidth,center}
\setlength{\tabcolsep}{4pt}
\begin{tabular}{llccc}\hline
Domain & Task & \textbf{Train} & \textbf{Test} & \textbf{Bench}\\ \cline{1-5}
\multirow{3}{*}{Image} &Layout Planning& 27,246 & 1,255 & --\\
& Keywords Aug. & 22,364 & 1,039 & --\\
& Layout Shift & 22,364 & -- & --\\ \hline
\multirow{2}{*}{Text} & Layout Planning & 4,873 & -- &  502 \\
 & Text Split & 4,873 & -- &  --\\ \hline
\multirow{2}{*}{Graphic} & Layout Planning & 16119 & 1954 & -- \\
&  Layout Shift & 16176 & -- & -- \\
% \hline
\hline
\end{tabular}
\end{adjustbox}
\caption{
Number of examples in each training and testing of the InstructLap datasets. Tasks refer to the original layout planning data processed and its different augmentation methods.
% \roy{Text Split Instruction and Layout Shift Instruction are not clear.
}
\vspace{-2em}
\label{tab:data_statistics}
\end{table}

\subsection{MS COCO Layout}
We conducted data processing and filtering to remove abnormal samples, resulting in 27,246 text-to-layout pairs for layout planning tasks. We then generated textual descriptions for these layouts by prompting GPT-4V. we use different augmentation methods to obtain more instruction tuning data and split the dataset into training and testing for visual layout planning task. 
% The details are provided below.

\vspace{-0.5em}
\paragraph{Layout Normalization}
We standardized the size of the images in MS COCO by scaling the longest axis to a uniform length and centering it within a square canvas. This adjustment also involved recalibrating the bounding box coordinates to ensure that the scaling process did not distort the appearance of objects.
\vspace{-0.5em}
\paragraph{Object Selection}
In the annotations, small bounding boxes were discarded to simplify the layout. An area threshold $\tau_a=0.1$ was applied to the largest box in an image to avoid layouts cluttered with numerous small objects. Furthermore, a threshold $\tau_l=0.2$ was established for the maximum dimension (height or width) of all boxes in each image to remove background objects that are too small or irrelevant to the overall scene. These thresholds were manually selected.
\vspace{-0.5em}
\paragraph{Overlapping Filtering}
Overlapping bounding boxes were considered undesirable as they could introduce conflicting guidance in the same region, potentially leading to localized inaccuracies or ``hallucinations'' in the generated image. To address this issue, we calculated a pairwise matrix of Intersection over Union (IoU) scores for all bounding boxes within each image. A threshold $\tau_o=0.01$ was selected as the maximum value to filter out images with significant overlaps of the boxes.
\vspace{-0.5em}
\paragraph{Crowd Filtering}
Images in MS COCO with the `is\_crowd' label in their annotations were removed. This label indicates cases where a bounding box encompasses multiple instances of the same class under a single label (\textit{e.g.}, a densely populated area of people annotated with the singular label "person"). Such cases can potentially make LLMs confused when estimating the size of an individual.

\subsection{Graphic Design Data}
We also built InstLap based on Crello~\cite{yamaguchi2021canvasvae} for automatic graphic designs. Specifically, we first merge three types of elements into the background: i) tiny elements that occupy less than 1\% of the canvas and ii) elements with more than 70\% transparent pixels, which means that their shapes are difficult to describe as bounding boxes. We then filter the dataset based on the number of remaining elements, discarding those with more than 10 elements resulting in 16,119 training layouts and 1,954 layouts for testing. 

\subsection{LLM-Assisted Annotation}

\paragraph{Layout Prompt Generation}
We observed a significant inconsistency between object detection and caption annotations within the MS COCO dataset. Frequently, many objects either lack references in any of the five captions or are described using different terminology. For example, an object labeled as "person" in the annotation might be referred to as "woman" in the caption. As a result, caption annotations cannot reliably serve as prompts for layout generation. A detailed example is given in Appendix \ref{sec:caption_example}.

To address this challenge, we capitalize on the observational and interpretative capabilities of large vision-language models, such as GPT-4V ~\citep{yang2023dawn}, LLaVA ~\citep{liu2023llava}, and Phi-3-Vision ~\citep{abdin2024phi}. We utilize these models to generate text prompts that describe given layouts, which consist of selected objects based on box annotations and the original image. To ensure consistency between the generated prompt and layout, we provide manually crafted examples to illustrate desired captions for specific layouts. Additionally, we instruct the model not to mention any objects in the image that are not included among the selected objects.

In the Crello dataset, the elements are too varied to be categorized or described using a predetermined set of labels. Consequently, we utilized the Phi-3-Vision model to create a brief description for every element in the dataset, including the background for content-aware graphic designs. In addition, we employed a heuristic method to create spatial descriptions for major elements in the design. All element captions, sizes, and spatial instructions are added in the prompt, providing contexts for LLM when generating vibrant layouts.

\paragraph{Object Augmentation}
The MS COCO dataset employs a set of labels of 80 classes, which would restrict the generalizability of LLM models. We utilize an LLM to augment the labels to facilitate open-set layout inference. Specifically, we provide carefully curated examples as context for GPT-4 \citep{achiam2023gpt} and ask it to replace the single word tag of a box with a phrase that describes an object of comparable size and similar semantics, ensuring that the enhancement remains concise and natural.

\paragraph{Layout Shift Augmentation}
We enrich the training data with instruction-following pairs to promote the model's understanding of the relationship between bounding-box coordinates and directions. We guide the model in moving an object in a specified direction for a random distance. To prevent overlaps, we ask the model to move objects at the boundary further from adjacent objects, {\it e.g.}, moving the leftmost object further to the left. This strategy boosts the model's spatial awareness and ensures the integrity of the data. 

\subsection{Visual Text Layout Data}
We have curated a collection of 4,873 layout prompts specifically tailored for visual text analysis. These prompts were generated from TextDiffuser-2 training data samples \cite{chen2023textdiffuser2}, which include caption-OCR pairs derived from the MARIO-10M dataset. We developed two distinct prompts for each layout to improve TextLap's ability to adhere to user instructions for keyword splitting and to autonomously handle keyword segmentation. One prompt explicitly demonstrates how keywords are split, providing clear guidance for the process. The other prompt delegates the segmentation task to TextLap, allowing the model to autonomously determine the optimal segmentation approach based on the layout context.

\paragraph{InstaLap Bench}
We build InstLap Bench following the TRINS-Gen dataset~\cite{trins2024}.
Considering the difficulty of rendering too many words in a single image, we run Paddle OCR and filter out images with more than 10 OCR words, resulting in a curated set of 502 images. We hired annotators from Upwork to write human annotations for each given image, which describe both textual and visual objects. Each annotator has to explicitly describe words and their positions within the image. To this end, we introduce InstLap-Bench, the visually-rich document design benchmark meticulously annotated by humans.

\section{Model Training}

We construct the TextLap model by fine-tuning the Vicuna-1.5 7B \citep{vicuna2023} model using the FastChat framework \citep{zheng2024judging}. The model is trained on the InstLap-train set with both object layouts and text layouts. Figure \ref{fig:InstLap_overview} presents examples of conversations from the InstLap-train dataset. We train the model using both the closed-set layout prompt and the prompt enhanced with object augmentations. The training aims to enable the model to extract objects and separate visual-text elements when no specific target is provided. Additionally, the model is trained to focus on user-specified keywords or modification instructions. Training details are included in Appendix \ref{sec:train_detail}.

We use multiple text formats for layout representation, including lists of integer/float numbers and CSS structures. However, we notice that multi-digit coordinates are often tokenized into multiple tokens, potentially complicating the model's numerical understanding. We integrated discrete coordinates as single tokens into the tokenizer to mitigate this. 
\section{Experimental Results}

\subsection{Experiments Setup}
We evaluate our model’s capability to generate layouts for objects and visual text using the InstLap-Test split. All TextLap varaints utilize the same set of weights trained on the InstLap-Train set. We first assess the model’s performance in close-set object layout generation, which feature a defined label set of 80 classes, and an open-set scenario that include LLM-augmented object descriptions. In both experiments, we present prompts with and without specific objects to evaluate the TextLap on identifying key objects within the text prompts. 

Then, we conduct an experiment using the Crello dataset \cite{yamaguchi2021canvasvae} for graphic design given a list of elements, including images, texts, and SVGs.

In addition, we evaluate our model for identifying visual-text elements and creating visual-text layouts using the InstLap-Bench split. The prompts in this dataset are significantly more complex than the 5k visual-text training samples in the InstLap-Train set. All experiments are implemented with PyTorch and performed on Nvidia A100 GPUs.

\subsection{Visual Layout Generation}
While image generation has greatly benefited from the continual expansion of diffusion models, controllability remains a challenge, particularly in scenarios involving multiple user-imposed conditions or constraints. Current generative models like DALL·E 3 \citep{betker2023improving} and SDXL \citep{podell2023sdxl} still struggle to generate samples that satisfy all conditions. These challenges underscore the limitations of relying solely on larger models, prompting the need to incorporate additional layout planning components to better address the problem.
\vspace{-1.5em}
\paragraph{Evaluation Metrics}
To assess the quality of the generated content, we employ six computational metrics: (1) Fréchet Inception Distance (FID) \citep{heusel2017gans}: Assesses the similarity between distributions of generated layouts and validated layouts, capturing both the variety and fidelity of the generated layouts. We modified the network structure to adapt the open-set layout. The model is introduced in Appendix \ref{sec:LFID}. (2) MaxIoU (Maximum Intersection-over-Union): this metric assesses the overlap between the generated and target layouts by calculating the average IoU of the optimally matched element pairs. We adapt the original definition by \cite{kikuchi2021constrained}, focusing on matching layouts generated from the same prompt rather than from the same set of labels. For object layouts, we find the best-matched box among the boxes that share the same label. Finding exact matches is challenging for visual-text layouts, where box labels are segments of phrases that may be split differently. Therefore, we define the closest match using the cosine similarity of CLIP text features. (3) Failure rate: the proportion of layouts generated by the LLM in an invalid format that cannot be processed automatically. (4) Precision: Accuracy of correctly identified elements within all extracted elements. (5) Recall: The ratio of correctly identified elements within all ground truth elements. (6) F-score: The harmonic mean of Precision and Recall, balancing their contributions.
\begin{table}[!ht]
\centering
\fontsize{6}{7.5}\selectfont
\resizebox{1.0\linewidth}{!}{
\setlength{\tabcolsep}{2pt}
\vspace{-1.5em}
\begin{tabular}{lccccc}
\hline
\multirow{2}{*}{Methods} & \multicolumn{4}{c}{Text prompts with target objects} \\ \cline{2-5} 
 & FID ↓ & MaxIoU ↑ & Fail \% ↓  & F-score ↑ \\ \hline
GPT-4 & 382.0 & 0.292 & 0.956  & 0.989 \\
GPT-4 (R) & 26.40 & 0.452 & 1.116  & 0.979 \\
GPT-4 (rCSS) & 37.45 & 0.459 & 1.116  & 0.969 \\ 
LayoutGPT & 248.0 & 0.435 & {\textbf{0.000}} & 0.959 \\ \hline
TextLap-S128 & 18.64 & 0.454 & 2.151 &  0.974 \\
TextLap-D128 & {\textbf{13.54}} & {\textbf{0.475}} & 0.398 &  0.983 \\
TextLap-D1024 & 14.43 & 0.456 & 0.398 &  0.973 \\
TextLap-Float & 15.09 & {\textbf{0.475}} & 0.159 &  0.996 \\ 
TextLap-CSS & 19.32 & 0.458 & {\textbf{0.000}} & {\textbf{0.998}} \\ \hline
 & \multicolumn{4}{c}{Text prompts only} \\ \hline
GPT-4 & 504.0 & 0.005 & 98.566  & 0.012 \\
GPT-4 (R) & 30.01 & 0.417 & 4.382 & 0.856 \\
GPT-4 (rCSS) & 39.17 & 0.427 & 1.594  & 0.867 \\ 
LayoutGPT & 265.7 & 0.416 & {\textbf{0.000}} & 0.900 \\ \hline
TextLap-S128 & 20.67 & {\textbf{0.452}} & 1.116  & 0.976 \\
TextLap-D128 & 14.77 & 0.267 & 25.976  & 0.537 \\
TextLap-D1024 & 16.27 & 0.347 & 9.163  & 0.716 \\
TextLap-Float & {\textbf{14.36}} & 0.424 & 3.426  & 0.877 \\ 
TextLap-CSS & 18.69 & 0.440 & 0.080 & {\textbf{0.979}} \\ \hline
\end{tabular}
}
\vspace{-1em}
\caption{Comparative results of close-set layout generation with 80-class COCO labels}
\vspace{-1.5em}
\label{tab:Close-set-table}
\end{table}

\begin{table}[!ht]
\centering
\fontsize{6}{7.5}\selectfont
\resizebox{1.0\linewidth}{!}{
\setlength{\tabcolsep}{2pt}
\begin{tabular}{lccccc}
\hline
\multirow{2}{*}{Methods} & \multicolumn{4}{c}{Text prompts with target objects} \\ \cline{2-5} 
 & FID ↓ & MaxIoU ↑ & Fail \% ↓  & F-score ↑ \\ \hline
GPT-4 & 291.8 & 0.330 & 1.347  & 0.979 \\
GPT-4 (R) & 48.59 & 0.421 & 1.925  & 0.974 \\
GPT-4 (rCSS) & 41.73 & 0.456 & 0.481  & 0.972 \\ 
LayoutGPT & 261.4 & 0.383  & {\textbf{0.000}}  & 0.824 \\ \hline
TextLap-D128 & {\textbf{17.77}} & {\textbf{0.485}} & 1.059  & 0.972 \\
TextLap-D1024 & 21.18 & 0.467 & 0.385  & 0.949 \\
TextLap-Float & 18.16 & 0.456 & 0.481  & 0.937 \\  
TextLap-CSS & 19.48 & 0.464 & {\textbf{0.000}} & {\textbf{0.996}} \\ \hline
 & \multicolumn{4}{c}{Text prompts only} \\  \hline
GPT-4 & 398.4 & 0.006 & 95.091  & 0.011 \\
GPT-4 (R) & 78.03 & 0.155 & 9.047  & 0.315 \\
GPT-4 (rCSS) & 88.66 & 0.093 & 2.406  & 0.197 \\ 
LayoutGPT & 269.4 & 0.326 & {\textbf{0.000}} & 0.712 \\ \hline
TextLap-D128 & {\textbf{16.38}} & 0.322 & 18.383  & 0.642 \\
TextLap-D1024 & 18.34 & 0.349 & 10.298  & 0.714 \\
TextLap-Float & 16.67 & 0.309 & 3.272  & 0.633 \\ 
TextLap-CSS & 19.44 & {\textbf{0.446}} & 0.192 & {\textbf{0.960}} \\ \hline
\end{tabular}
}
\caption{Results on open-set layout generation with LLM-augmented labels}
\vspace{-3em}
\label{tab:Open-set-table}
\end{table}

\vspace{-0.5em}
\paragraph{Settings}
To evaluate the effectiveness of text-to-layout models, we use text-to-image generation as a downstream task. We compare our method, TextLap, with LayoutGPT \cite{feng2023layoutgpt} and GPT-4-based baselines for close-set and open-set text-guided layout generation. The naive baseline is termed GPT-4, where we ask GPT-4 to generate a layout for a given prompt with three fixed examples to demonstrate the desired output format. We introduce two additional baselines, GPT-4 (R) and GPT-4 (rCSS), for our experiments designed to enhance in-context learning for the GPT-4 model. These baselines employ cosine similarity of CLIP text features to identify and retrieve the most relevant demonstration examples from the InstLap-Train dataset for each test prompt~\cite{feng2023layoutgpt}. They differ in their approach to representing layout coordinates: GPT-4 (R) uses lists of integers, while GPT-4 (rCSS) adopts a CSS-like structure with a maximum value of 128. We finetuned four TextLap variants with different coordinate representations: discrete coordinates on 128×128 and 1024×1024 canvases (TextLap-D128, TextLap-D1024), floating-point coordinates (TextLap-Float), and a CSS format with a max value of 128 (TextLap-CSS). Additionally, we trained a model with 128 new tokens for discrete coordinates, called TextLap-S128.

% \vspace{-0.5em}
\paragraph{RQ1: Does simply scale up the LLM help for design problems?}
Table \ref{tab:Close-set-table} presents the results of TextLap models alongside three GPT-4-based methods. In particular, TextLap models achieve significantly lower layout FID scores compared to GPT-4 baselines, highlighting the advantages of fine-tuning. In the open-set generation results, presented in Table \ref{tab:Open-set-table}, there is a notable decrease in the retrieval scores for GPT-4 baselines when objects are not specified in the prompt. This decline could be attributed to the fact that, in close-set experiments, the retrieval process often identifies examples with matching objects, implicitly providing the necessary keywords in context. This shows that task-specific models can potentially outperform larger models and simple scaling-up does not help for design problems.

% \vspace{-0.5em}
\paragraph{RQ2: What is the impact of different layout representations?}
Among various TextLap variants, TextLap-CSS stands out, outperforming all baselines by a considerable margin and demonstrating high stability in object extraction. TextLap-S128 showed inferior performance, likely due to inadequate training data to effectively integrate new weights. GPT-4 (rCSS) also exhibits the lowest failure rates among the baseline methods, consistent with insights from previous studies by \cite{yang2024if, feng2023layoutgpt, tang2023layoutnuwa}, which suggest that pre-trained large language models better understand programming patterns, possibly due to exposure to code snippets in their training data.

% \vspace{-0.5em}
\paragraph{RQ3: Are special tokens of coordinates useful?} TextLap-S128, which incorporates special coordinate tokens, performs well on closed-set data, with a lower failure rate in text prompt-only settings compared to TextLap-D128 and TextLap-Float. However, it struggles in open-set generation, likely due to the large number of new parameters requiring more training data and time. 
\begin{figure}[!ht]
    \centering
    \includegraphics[width=1\linewidth]{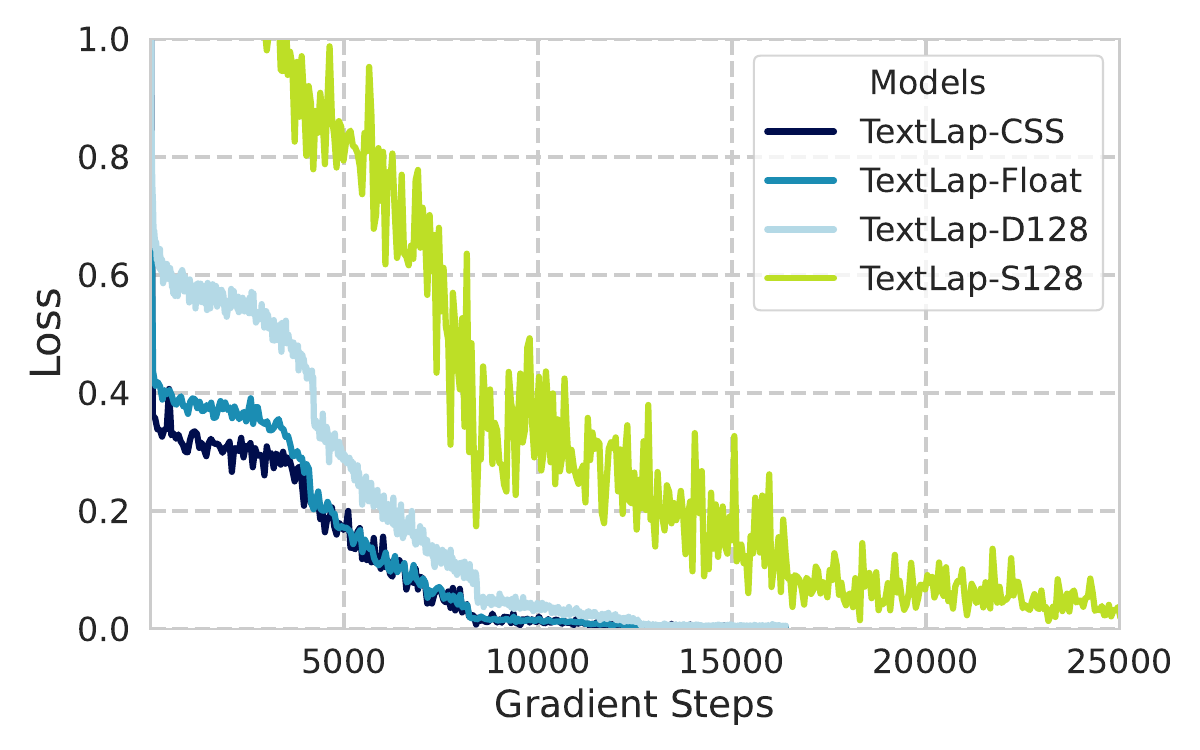}
    % \vspace{-1em}
    \caption{Loss curve on close-set layout generation with 80-class COCO labels.}
    \label{fig:loss}
\end{figure}
As shown in the loss curve in Figure \ref{fig:loss}, TextLap-S128 had higher loss after the same number of epochs and converged to a slightly higher loss even with extended training, indicating a need for more training to achieve similar performance. Fine-tuning existing tokens, which are well-trained in LLMs and handle arithmetic effectively, proves more efficient. Thus, adding special tokens is less effective than a CSS-based layout representation, especially for open-set tasks, where large-scale pre-training may help~\cite{lv2023kosmos}.

\subsection{Visual Text Layout Generation}
We evaluate the ability of our models in generating layouts for visual-text using the InstLap-Bench set. Specifically, we compare TextLap-Float and TextLap-CSS with GPT-4 and GPT-4 (rCSS). Additionally, we benchmark our models against the large language model (LLM) employed in TextDiffuser2, which is fine-tuned on the same 5,000 visual-text layout training samples, utilizing the Vicuna-v1.5 7B checkpoint. 

\begin{table}[h!]
\centering
\fontsize{6}{7.5}\selectfont
\resizebox{\linewidth}{!}{
\setlength{\tabcolsep}{4pt}
\begin{tabular}{lccc}
\hline
Methods & MaxIoU (suc) ↑ & MaxIoU ↑ & Fail \% ↓ \\ \hline
GPT-4 & 0.231 & 0.206 & 10.778 \\
GPT-4 (rCSS) & 0.252 & 0.241 & 4.192 \\ 
TextDiffuser-2 & 0.166 & 0.165 & 0.80 \\ \hline
TextLap-Float & 0.209 & 0.096 & 54.09 \\ 
TextLap-CSS & 0.211 & 0.211 & 0.00 \\ \hline
\end{tabular}}
\caption{Results on text layout generation on the InstLap-Bench.}
\label{tab:InstLap-Bench}
\end{table}

Table \ref{tab:InstLap-Bench} shows the results comparion between TextLap and GPT-4. Given the high failure rates of GPT-4 and TextLap-Float, we also present MaxIoU (suc), which represents the average MaxIoU across all successful generations. In particular, TextLap-CSS achieves the lowest failure rate and significantly outperforms TextDiffuser2 in MaxIoU. However, it fails to outperform the GPT-4 baselines. This limitation is likely attributed to insufficient training data, stemming from the limited number of visual-text samples in the InstLap-train dataset and the inconsistent distribution of layouts between InstLap-train and InstLap-Bench.

\subsection{Automatic Graphic Designs}
\begin{figure}[!ht]
    \centering
    \includegraphics[width=1\linewidth]{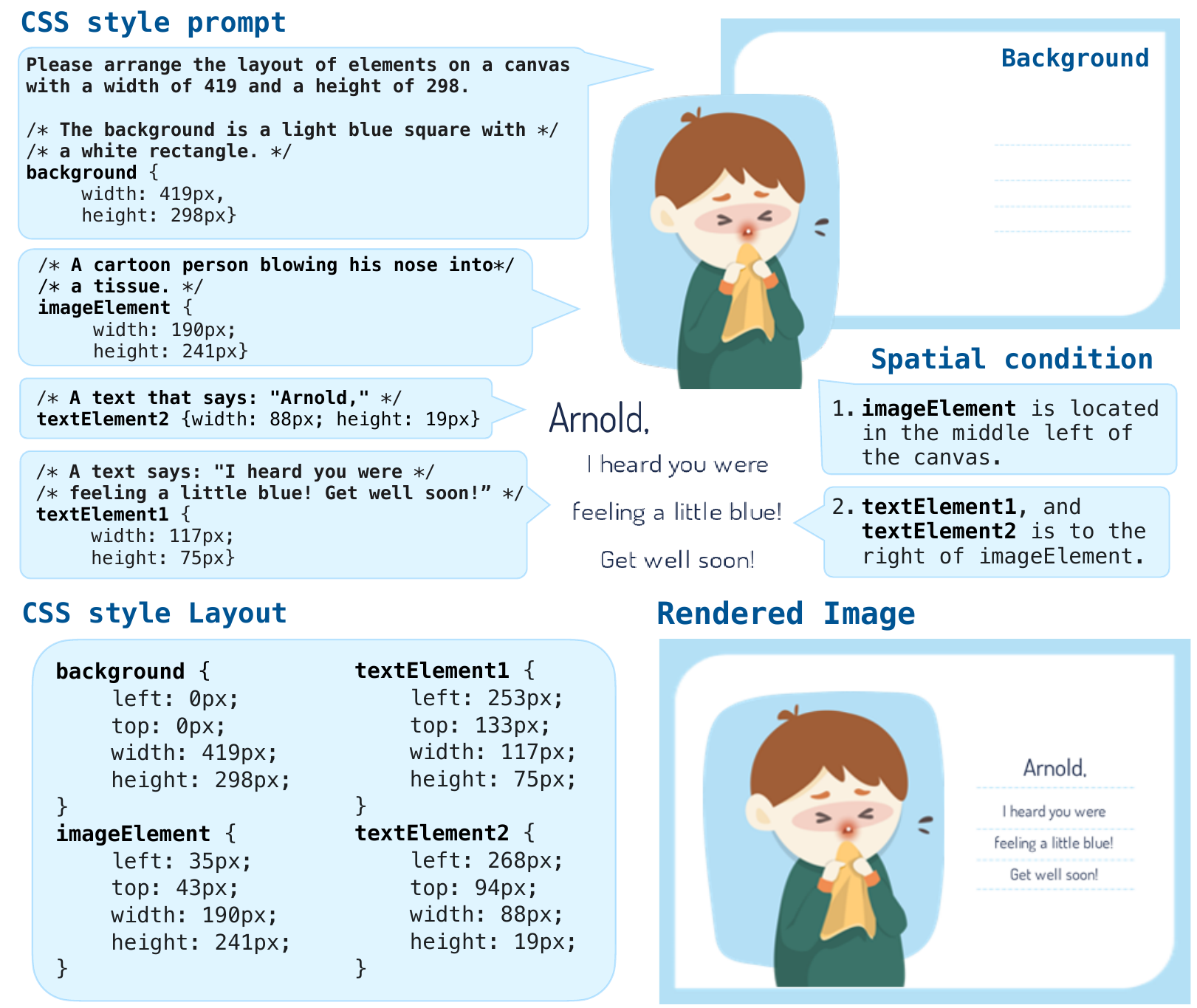}
    \vspace{-1em}
    \caption{An example from InstLap that is built based on the Crello dataset.}
    \label{fig:crello}
\end{figure}

We conducted an experiment on content-aware graphic designs using the Crello dataset, where canvas sizes varies for each design. For CSS format layouts with integer coordinates, we include the canvas size in the prompt. We also normalize box coordinates to decimal values between 0 and 1 ("Float") and write layouts as JSON to leverage the model's code understanding. Figure \ref{fig:crello} illustrates a generated example using CSS forat, where TextLap provides coordinates for each element given a text instruction and a list of elements.

\begin{table}[!h]
\centering
\fontsize{6}{7.5}\selectfont
\resizebox{\linewidth}{!}{
\setlength{\tabcolsep}{2pt}
\begin{tabular}{lccccc}
\hline
 & MaxIoU↑ & Precision↑ & Recall↑ & F-score↑ \\ \hline
GPT-4 (CSS) & 0.190 & 0.874 & 0.257 & 0.364 \\
GPT-4 (rCSS)  & 0.209 & 0.934, & 0.296, & 0.406 \\
GPT-4 (Float) & 0.457 & 0.980 & 0.980 & 0.980 \\
GPT-4 (rFloat)  & 0.440 & 0.984 & 0.984 & 0.984\\
\hline
TextLap-CSS & 0.407 & 0.998 & 0.986 & 0.990  \\
TextLap-Float & 0.535 & 1.000 & 1.000 & 1.000 \\
\hline
\end{tabular}
}
\caption{Results on automatic graphic design on Crello.}
% \vspace{-1em}
\label{tab:graphicdesign}
\end{table}

We compare TextLap with four GPT-4 baselines that generate CSS and JSON-style layouts with both integer and float coordinates. SBERT retrieves demonstration layouts for the GPT-4 (rCSS, rFloat) baselines. Table \ref{tab:graphicdesign}, shows TextLap significantly outperforms its GPT-4 counterpart, demonstrating the effectiveness of instruction fine-tuning. Additionally, methods using float coordinates outperform those using integer coordinates, likely due to the complexity of adapting to different canvas sizes.

\begin{figure*}[!ht]
    \centering
\includegraphics[width=\textwidth]{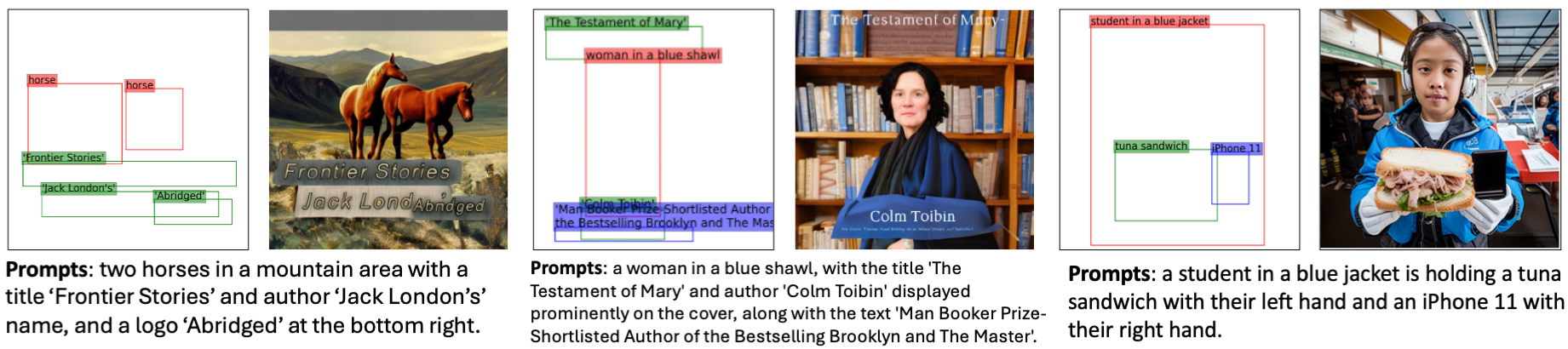}
    \vspace{-1.5em}
    \caption{\fontsize{9.5}{10}\selectfont Generated visual and textual layout planning examples. Layouts are provided by TextLap given text prompts and images are rendered by ARTIST~\cite{zhang2024artist} and InstanceDiffusion~\cite{wang2024instancediffusion} respectively.}
    \label{fig:vlplanning1}
\end{figure*}

\begin{figure*}[!hb]
    \centering
    \includegraphics[width=1\textwidth]{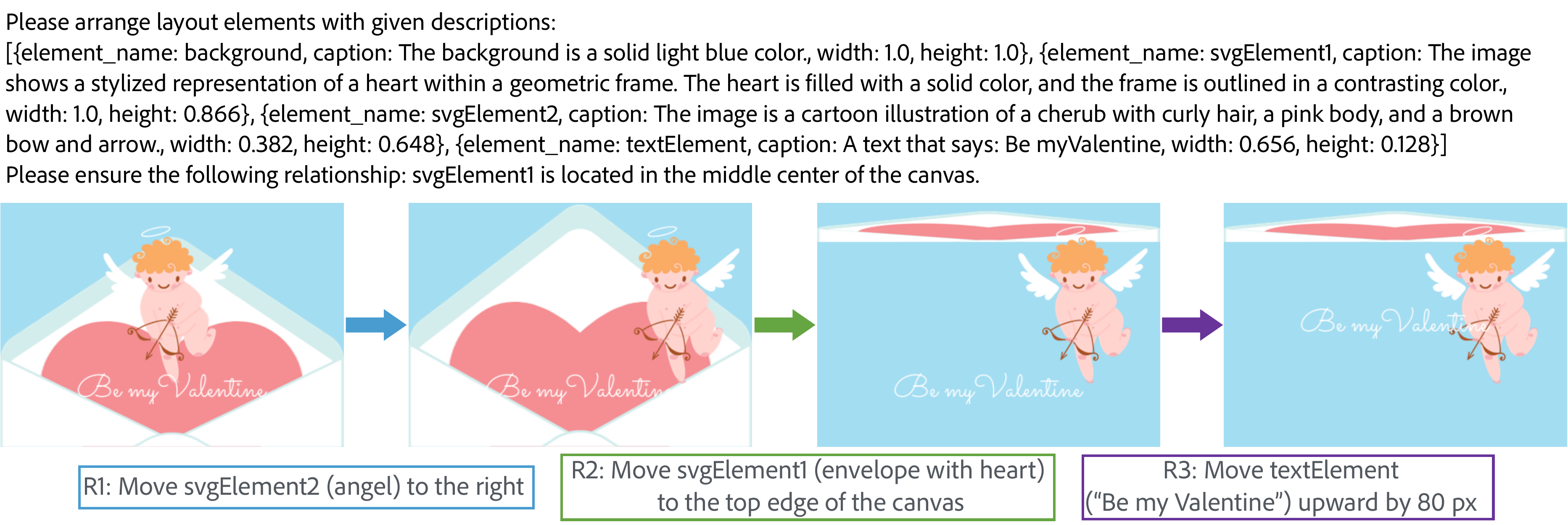}
    \caption{\fontsize{9.5}{10}\selectfont Examples of interactive layout design, where user gives instructions to TextLap.}
    \vspace{-1em}
    \label{fig:inter}
\end{figure*}

\subsection{Impact of Layout on Text-to-Image Generation}
We evaluated the effect of layout guidance on text-to-image generation using the COCO dataset. The experiment compared three models: Stable Diffusion 1.5, which serves as the layout-free baseline; TextLap + InstanceDiffusion \cite{wang2024instancediffusion}, where layouts are generated by our TextLap model to guide InstanceDiffusion; and True Layout + InstanceDiffusion, where the actual COCO layout is used for guidance. 
\begin{table}[!t]
\centering
\fontsize{6}{7.5}\selectfont
\resizebox{\linewidth}{!}{
\setlength{\tabcolsep}{4pt}
\begin{tabular}{lcc}
\hline
\textbf{Method} & \textbf{Image FID} & \textbf{GPT-Preference} \\
\hline
SD 1.5 & 58.09 & - \\
InstDiff (TextLap) & 58.92 & 72\% \\
InstDiff (True Layout) & 58.39 & 76\% \\ \hline
\end{tabular}}
\caption{Text-to-image generation results on the COCO dataset. Layout-guided models are compared to the layout-free baseline, Stable Diffusion (SD) 1.5.}
\label{tab:t2l2i}
\end{table}
The models were evaluated using two metrics. The first metric, Image FID, measures the quality of the generated images by comparing them to real images from the COCO dataset, with lower FID scores indicating better quality. The second metric, GPT-Preference, involves presenting three images and the text caption to GPT-4o:  two images generated by a test model and Stable Diffusion 1.5 using the same prompt, and the real image used to generate the text caption of the layout. GPT-4o compares the generated images to determine which has a more similar layout to the real image and better coherence with the text caption. The results, shown in Table~\ref{tab:t2l2i}, indicate that layout guidance slightly impacts FID but improves text-image alignment. TextLap + InstanceDiffusion is preferred by GPT-4o 72\% of the time, while the true layout model achieves 76\%.

\subsection{Qualitative Evaluation}
\paragraph{Examples of Object Layout Generation}
We compare generated layouts between GPT-4 (rCSS) and InstLap-CSS and further evaluate them based on images rendered by InstanceDiffusion \citep{wang2024instancediffusion}, as shown in Figure \ref{fig:visplan}. Both GPT-4 and InstLap can generate layouts that follow the constraints from the prompts. The advantages of InstLap include: (i) generated layouts can simplify the process of creating visually appealing images, indicating that these layouts are of high quality and more suitable for open-source rendering engines, as reflected in evaluation metrics such as MaxIOU; and (ii) InstLap offers customization based on user needs through instruction tuning, with a model size significantly smaller than GPT-4. Examples of text layout generation are provided in Appendix \ref{sec:moreexamples}.

\paragraph{Emergent Visual-Text Layout Generation}
As shown in Table \ref{tab:data_statistics}, TextLap has been fine-tuned for text layout designs in TextDiffuser-2~\citep{chen2023textdiffuser2} and visual object designs on InstLap based on MS COCO 2017~\citep{caesar2018coco}. It shows surprising generalization ability and emergent visual-text joint layout planning ability. The great generalizability comes from a carefully designed dataset building strategy as described in Section \ref{sec:data}. Figure \ref{fig:vlplanning1} shows examples of visual-text layout generation, and more examples can be found in Appendix \ref{sec:moreexamples}.

\paragraph{Interactive Layout Design}
Figure \ref{fig:inter} and \ref{fig:inter2} shows an example of an interactive layout design, where the user can comment on existing layout designs, and TextLap can generate a new layout, fulfilling user requests. This is another emergent ability, as TextLap is never fine-tuned on conversational data.

\section{Conclusion}
This study addresses the text-to-layout task by creating an instructional dataset built upon available resources with the assistance of GPT-4v to fine-tune a large language model for layout planning. The fine-tuned model, TextLap, outperforms GPT-4 in object layout planning and can generate layouts with both text and object when trained on text-only and object-only layouts. TextLap provides a framework to addressing real-world graphic design challenges by building instruction-following layout datasets. It is desired if the image rendering models can be finetuned or jointly trained with InstLap. However, it is beyond the scope of this paper as InstLap aims to unveiling the graphic design ability of large language models.

\section{Limitations}
The limitations of the paper are caused by the design of the dataset and the model architecture. InstLap is a dataset created with complex heuristics and the help of large language models. There is still a quality gap between InstLap and high-quality human annotations. However, InstLap is much cheaper to build and can serve as a pioneer dataset to quickly verify whether language models can perform automatic graphic designs. TextLap uses the standard LLM architecture, and a special design with 2D spatial embedding layers should provide better performance. It is very expensive to pretrain such a model and is beyond the scope of this paper.

\section{Ethics Statement}
Our paper introduces a new instruction tuning dataset, which acknowledges the potential ethical implications inherent in using large language models for such applications.We have taken comprehensive steps to ensure that our research adheres to the highest ethical standards, particularly with respect to data privacy and responsible use of AI. This ethics statement reflects our dedication to conducting responsible research and our commitment to advancing the field of AI in a manner that respects individual privacy rights and promotes the ethical use of technology.

\subsubsection*{Acknowledgments}
This work is partially supported by NSF AI Institute-2229873, NSF RI-2223292, NSF III-1747614, an Amazon research award, and an Adobe gift fund. Any opinions, findings and conclusions or recommendations expressed in this material are those of the author(s) and do not necessarily reflect the views of the National Science Foundation, the Institute of Education Sciences, or the U.S. Department of Education.

% \bibliography{ref}

\clearpage
\appendix

\onecolumn
\counterwithin{figure}{section}
\counterwithin{equation}{section}
\counterwithin{table}{section}

\section*{\fontsize{15}{13}\selectfont \centering Appendix}
\vspace{2em}
\section{Comparison between Annotations in COCO and InstLap}
\label{sec:caption_example}
The key objective behind the development of the InstLap dataset is to ensure consistency in spatial relationships and object naming between captions and bounding box annotations. This alignment is critical for accurately mapping visual elements to descriptive captions, which is particularly important for tasks involving layout understanding.

To highlight the difference between datasets, consider the following five human-generated captions from the COCO dataset:
\begin{itemize}
    \item "A picture of a dog laying on the ground."
    \item "Dog snoozing by a bike on the edge of a cobblestone street."
    \item "The white dog lays next to the bicycle on the sidewalk."
    \item "A white dog is sleeping on a street and a bicycle."
    \item "A puppy rests on the street next to a bicycle."
\end{itemize}

While these captions provide descriptive information, they lack the consistency between annotations. In contrast, the InstLap dataset provides coherent and detailed captions with precise layout annotations, enabling a tighter alignment between descriptions, spatial relationships, and object details:
\begin{itemize}
    \item \textbf{Close-set caption:} "A dog is lying on the ground on the right with a bicycle parked to the left." \\
    \textbf{Objects:} [bicycle, dog]
    \item \textbf{Open-set caption:} "A sleepy labrador is lying on the ground on the right with a cherry red bicycle parked to the left." \\
    \textbf{Objects:} [cherry red bicycle, sleepy labrador]
\end{itemize}

\section{Implementation Detail}
\subsection{Training Detail}
\label{sec:train_detail}
All models are trained using eight NVIDIA A100 80GB GPUs. We use the default configuration load from the pre-trained Vicuna model. In fine-tuning, we use a cosine annealing schedule with an initial learning rate of 2e-5 and a batch size of 32. This set of hyperparameters is adopted across all checkpoints. 

\subsection{Open-Set Layout Encoder for FID Score Calculation}
\label{sec:LFID}
The FID score evaluates the quality of generated data by measuring the distance between feature vectors of real and generated samples, which requests a feature encoder that captures the position and semantic details of layout elements in a single vector. Adapting from the close-set approach by \cite{kikuchi2021constrained}, we develop our feature encoder for open-set layouts. The model uses a transformer-based encoder-decoder backbone, which encodes a sequence of layout elements to a feature vector and uses the decoder to reconstruct the input sequence. The feature is also trained to discriminate between clean and noisy layouts. Specifically, each element $\mathbf{x}=[\mathbf{b}, \text{CLIP}(s)] $ is defined as a four dimension bounding box $\mathbf{b}$ consisting of the left, top, right, bottom coordinates and the CLIP \cite{radford2021learning} text feature of the phrase $s$ of the element. The model is trained by three losses: A binary cross-entropy loss as the discrimination loss and a reconstruction loss consisting of a mean cosine distance loss for CLIP features and a mean square error loss for bounding boxes.

\section{Additional Quantitative Results}
Table \ref{tab:Merged-set-table} presents the results that combine open-set and close-set prompts, where TextLap models exhibit similar performance. This consolidation of results further underscores the consistent performance of TextLap models, showing that smaller LLMs customized for specific tasks can beat huge general LLMs, such as GPT4V.
\begin{table*}[!h]
\centering
\fontsize{6}{7.5}\selectfont
\resizebox{0.9\textwidth}{!}{
\begin{tabular}{lcccccc}
\hline
\multirow{2}{*}{Testing Methods} & \multicolumn{6}{c}{Captions with target objects} \\ \cline{2-7} 
 & FID ↓ & MaxIoU ↑ & Fail \% ↓ & Precision ↑ & Recall ↑ & F-score ↑ \\ \hline
GPT-4 & 382.0 & 0.292 & 0.956 & 0.989 & 0.989 & 0.989 \\
GPT-4(r) & 26.40 & 0.452 & 1.116 & 0.979 & 0.980 & 0.979 \\
GPT-4(rCSS) & 37.45 & 0.459 & 1.116 & 0.969 & 0.971 & 0.969 \\ \hline
TextLap-S128 & 18.64 & 0.454 & 2.151 & 0.972 & 0.977 & 0.974 \\
TextLap-D128 & {\textbf{13.54}} & {\textbf{0.475}} & 0.398 & 0.995 & 0.979 & 0.983 \\
TextLap-D1024 & 14.43 & 0.456 & 0.398 & 0.993 & 0.966 & 0.973 \\
TextLap-F & 15.09 & {\textbf{0.475}} & 0.159 & 0.996 & 0.997 & 0.996 \\ 
TextLap-CSS & 19.32 & 0.458 & {\textbf{0.000}} & {\textbf{0.998}} & {\textbf{1.000}} & {\textbf{0.998}} \\ \hline
 & \multicolumn{6}{c}{Captions only} \\ \hline
GPT-4 & 504.0 & 0.005 & 98.566 & 0.012 & 0.012 & 0.012 \\
GPT-4(r) & 30.01 & 0.417 & 4.382 & 0.838 & 0.900 & 0.856 \\
GPT-4(rCSS) & 39.17 & 0.427 & 1.594 & 0.842 & 0.925 & 0.867 \\ \hline
TextLap-S128 & 20.67 & {\textbf{0.452}} & 1.116 & 0.977 & 0.980 & 0.976 \\
TextLap-D128 & 14.77 & 0.267 & 25.976 & 0.538 & 0.540 & 0.537 \\
TextLap-D1024 & 16.27 & 0.347 & 9.163 & 0.718 & 0.719 & 0.716 \\
TextLap-F & {\textbf{14.36}} & 0.424 & 3.426 & 0.878 & 0.882 & 0.877 \\ 
TextLap-CSS & 18.69 & 0.440 & {\textbf{0.080}} & {\textbf{0.983}} & {\textbf{0.981}} & {\textbf{0.979}} \\ \hline
\end{tabular}
}
\caption{Comparative results of close-set layout generation with 80-class COCO labels}
\label{tab:Close-set-table_full}
\end{table*}

\begin{table*}[!h]
\centering
\fontsize{6}{7.5}\selectfont
\resizebox{0.9\textwidth}{!}{
\begin{tabular}{lcccccc}
\hline
\multirow{2}{*}{Testing Methods} & \multicolumn{6}{c}{Caption with target objects} \\ \cline{2-7} 
 & FID ↓ & MaxIoU ↑ & Fail \% ↓ & Precision ↑ & Recall ↑ & F-score ↑ \\ \hline
GPT-4 & 334.3 & 0.309 & 1.133 & 0.985 & 0.984 & 0.984 \\
GPT-4(r) & 27.99 & 0.438 & 1.482 & 0.977 & 0.978 & 0.977 \\
GPT-4(rCSS) & 32.54 & 0.458 & 0.828 & 0.970 & 0.972 & 0.971 \\ \hdashline
TextLap-D128 &  {\textbf{8.49}} &  {\textbf{0.479}} & 0.697 & 0.991 & 0.973 & 0.978 \\
TextLap-D1024 & 10.52 & 0.461 & 0.392 & 0.983 & 0.953 & 0.962 \\
TextLap-F & 9.96 & 0.466 & 0.305 & 0.969 & 0.970 & 0.969 \\
TextLap-CSS & 12.53 & 0.461 &  {\textbf{0.000}} &  {\textbf{0.997}} &  {\textbf{0.999}} &  {\textbf{0.997}} \\ \hline
\multicolumn{1}{l}{} & \multicolumn{6}{c}{Caption only} \\ \hline
GPT-4 & 394.1 & 0.005 & 96.992 & 0.011 & 0.013 & 0.011 \\
GPT-4(r) & 43.07 & 0.298 & 6.495 & 0.589 & 0.664 & 0.611 \\ 
GPT-4(rCSS) & 53.38 & 0.276 & 1.962 & 0.543 & 0.610 & 0.563 \\ \hdashline
TextLap-D128 &  {\textbf{8.46}} & 0.292 & 22.537 & 0.587 & 0.587 & 0.585 \\
TextLap-D1024 & 10.75 & 0.348 & 9.677 & 0.718 & 0.717 & 0.715 \\
TextLap-F & 9.74 & 0.372 & 3.357 & 0.768 & 0.771 & 0.767 \\
TextLap-CSS & 12.46 &  {\textbf{0.443}} &  {\textbf{0.131}} &  {\textbf{0.974}} &  {\textbf{0.972}} &  {\textbf{0.970}} \\ \hline
\end{tabular}
}
\caption{Comparative results on layout generation with both close-set and open-set labels.}
\label{tab:Merged-set-table}
\end{table*}

\clearpage
\section{Qualitative Comparison with GPT-4 Results}
\begin{figure*}[!h]
    \centering
    \includegraphics[width=\linewidth]{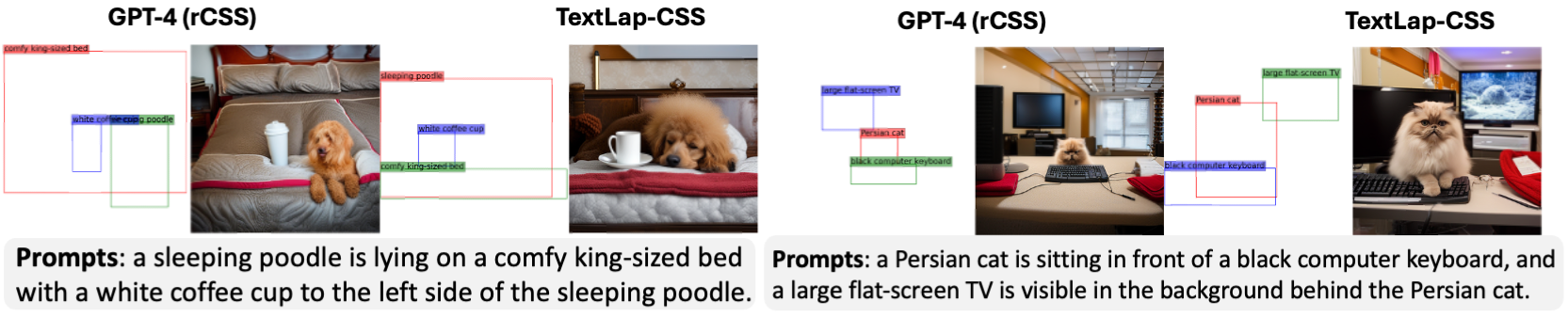}
    \caption{\fontsize{11}{10}\selectfont Comparison between TextLap-CSS and GPT-4 (rCSS). Images are rendered by InstanceDiffusion~\cite{wang2024instancediffusion} given layouts.} 
    \label{fig:visplan}
\end{figure*}

\section{Examples of Interactive Editing}
\label{sec:moreinter}
\vspace{-1em}
\begin{figure*}[!htp]
    \centering
    \includegraphics[width=1\textwidth]{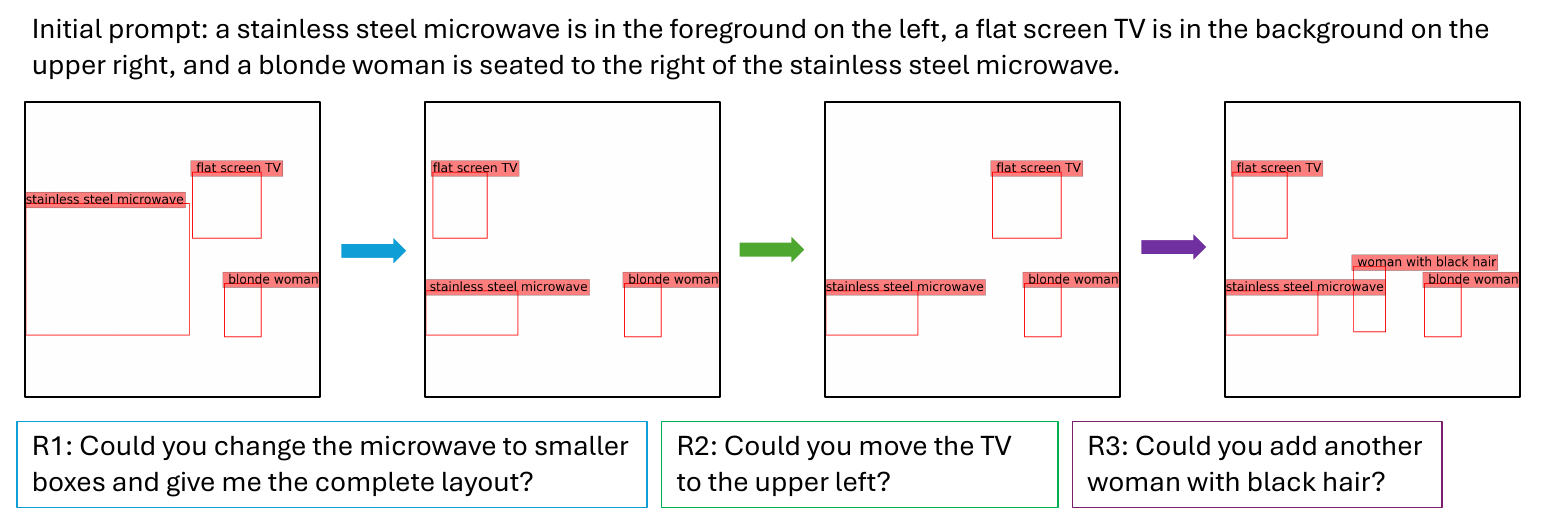}
    \vspace{-2em}
    \caption{\fontsize{11}{10}\selectfont Examples of interactive layout design on COCO dataset, where user gives instructions to TextLap.}
    \vspace{-1em}
    \label{fig:inter2}
\end{figure*}

\section{Examples of Text Layout Generation} 
\label{sec:moreexamples}
\begin{figure*}[!ht]
    \centering
    \includegraphics[width=\textwidth]{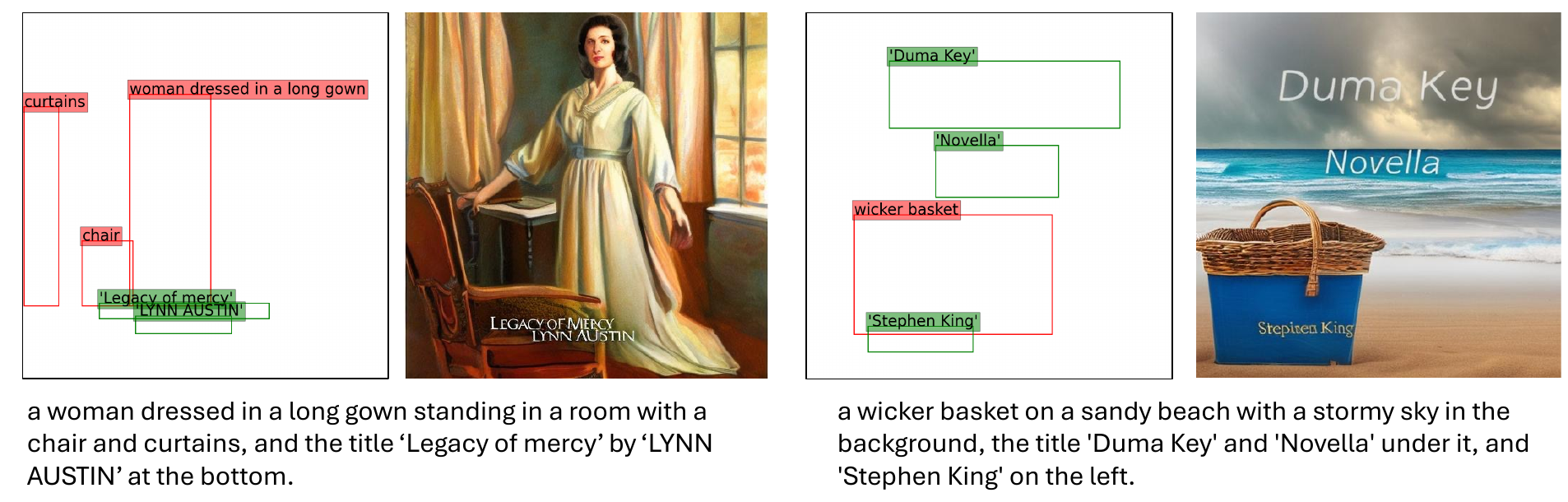}
    \caption{\fontsize{9.5}{10}\selectfont Examples of generated layout including both visual and textual elements by TextLap. Green boxes are text elements and Red boxes are visual elements of the generated layouts.}
    \label{fig:vlplanning}
\end{figure*}

Figure \ref{fig:vlplanning} shows examples of text and visual layout generation, where InstCap can detect keywords, objects and plan their locations within the image.

\clearpage
\begin{figure*}[!ht]
    \centering
    \includegraphics[width=\textwidth]{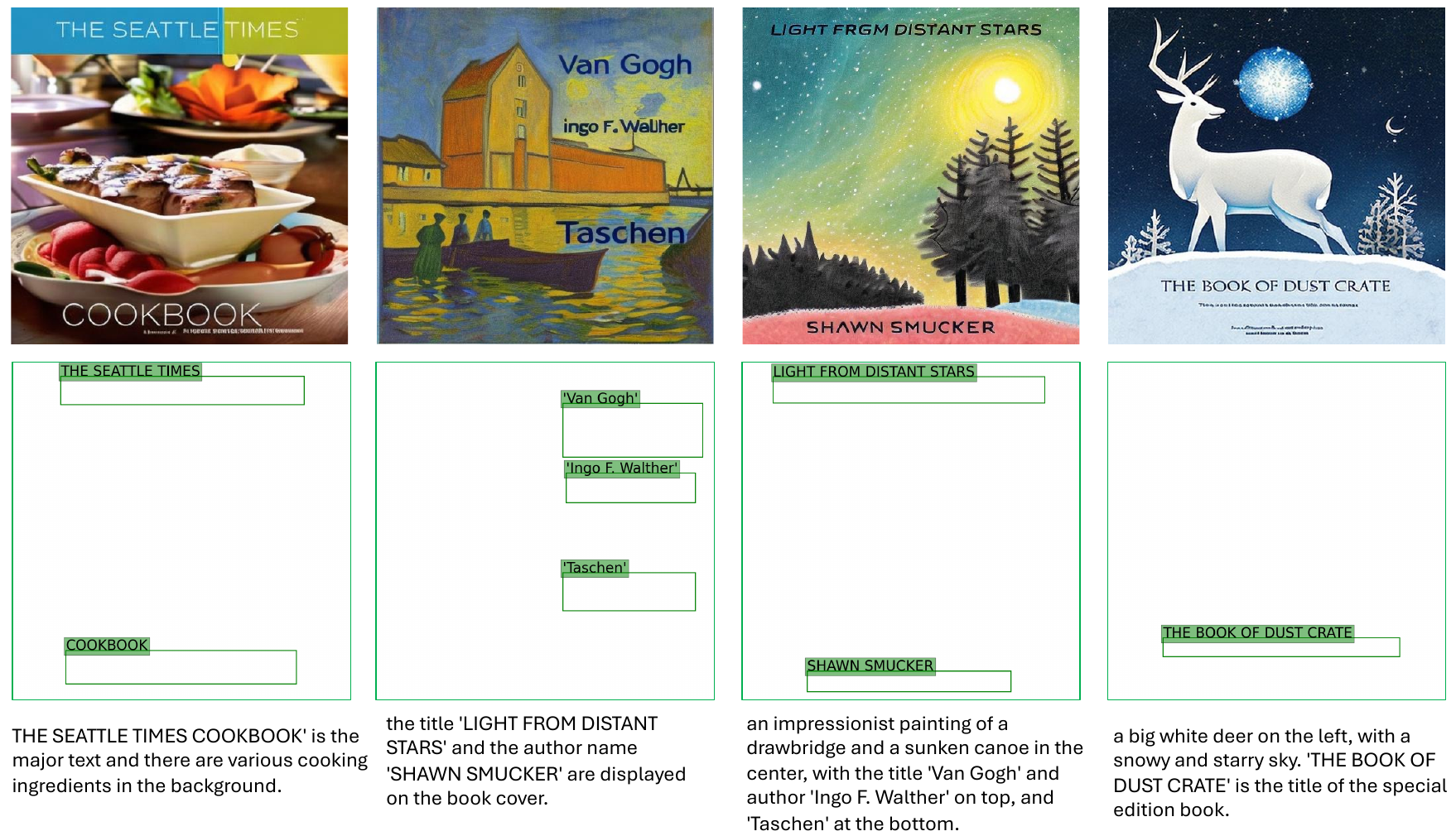}
    \vspace{-2em}
    \caption{\fontsize{9.5}{10}\selectfont Examples of designing text layouts where visual elements are generated based on the prompts.}
    \label{fig:textplan}
\end{figure*}

\end{document}